\pdfoutput=1

\documentclass[11pt]{article}

\usepackage{EMNLP2023}

\usepackage{times}
\usepackage{latexsym}

\usepackage[T1]{fontenc}

\usepackage[utf8]{inputenc}

\usepackage{microtype}

\usepackage{inconsolata}

\usepackage{amssymb}
\usepackage{pifont}
\usepackage{makecell}
\usepackage{multirow}
\usepackage{graphicx}
\usepackage{amsmath}
\usepackage{subfigure}
\usepackage[normalem]{ulem}
\useunder{\uline}{\ul}{}

\newcommand{\secref}[1]{Sec. \ref{#1}}
\newcommand{\figref}[1]{Figure \ref{#1}}
\newcommand{\eqnref}[1]{Eq. (\ref{#1})}

\title{Semantic Space Grounded Weighted Decoding for Multi-Attribute 
Controllable Dialogue Generation}

\author{Zhiling Zhang \hspace{5mm} Mengyue Wu$^*$\\
Shanghai Jiao Tong University \\
Shanghai, China\\
\texttt{\{blmoistawinde,mengyuewu\}@sjtu.edu.cn}
\And
Kenny Q. Zhu\thanks{~~Corresponding Authors.}\\
University of Texas at Arlington\\
Arlington, Texas, USA\\
\texttt{kenny.zhu@uta.edu}}

\begin{document}
\maketitle
\begin{abstract}
Controlling chatbot utterance generation with multiple attributes such as 
personalities, emotions and dialogue acts is a practically useful but 
under-studied problem.
We propose a novel framework called DASC 
that possesses strong controllability with a weighted decoding paradigm, 
while improving generation quality with the grounding in an 
attribute semantics space. Generation with multiple attributes is then 
intuitively implemented with an interpolation of multiple attribute embeddings,
which results in substantial reduction in the model sizes. 
Experiments show that DASC can achieve high control accuracy 
in generation task with the simultaneous control of 3 aspects while also producing interesting and 
reasonably sensible responses, even in an out-of-distribution robustness 
test. 
\footnote{Code and data are available at \url{https://github.com/blmoistawinde/DASC}.}
\end{abstract}

\section{Introduction}
\label{sec:intro}
Personalized dialogue systems are promising NLP applications for human-computer interaction and emotional companionship. We would expect such systems to have personalities, exhibit emotions, take dialogue acts and even adopt sophisticated strategies \citep{liu2021towards}, which necessitates the research efforts on \textit{Controllable Text Generation}. Despite recent progress in this field \citep{dathathri2019plug,keskar2019ctrl,krause2021gedi}, they mainly tackle single-attribute control, overlooking the fact that human interaction can usually convey multiple attributes simultaneously. Therefore, we explore a novel task of \textit{Multi-Attribute Controllable 
Dialogue Generation}, which can significantly ameliorate the expressiveness, human-likeness, and explainability of chat-bots. However, the numerous combinations of attributes can make the available data for each setting scarce, which poses a great challenge for this task.

Among previous works, \textit{Weighted Decoding} methods has achieved great success in single-attribute control tasks~\citep{arora2022director,liu-etal-2022-length}. 
Weighted decoding methods learn a token-level attribute classifier, 
which predicts the probability of the text conveying the desired attribute 
given the generation of each token in the vocabulary. 
Then the predicted probabilities are used to 
re-weigh the token generation during decoding to induce the attribute. 
Despite success in single-attribute 
control, they have certain limitations when extended to the multi-attribute case 
by multiplying several attribute predictions from multiple classifiers. Extra 
parameters proportional to the large vocabulary size $|V|$ will be introduced, 
which can grow several times further due to the number of attributes. 
The consequent large number of parameters will not only make the model inefficient, but also harm the generation quality. The model can be prone to overfit since the data for each attribute combination are usually small, which increases the risk of degeneration \citep{holtzman2019curious}. 

To overcome these limitations, we propose \textbf{D}ialog \textbf{A}ttribute \textbf{S}pace \textbf{C}ontroller (\textbf{DASC}). We establish an attribute semantic space where each token in the vocabulary is projected to the space through \textit{Attribute Token Embedding} shared across attributes. The language models' hidden states are also converted to \textit{Attribute Context Embedding} in the space through attribute-specific layers. The attribute space will be trained to make the tokens suitable to convey the desired attribute close to the current context embedding. We can then assign higher weights for the those tokens during decoding.
We will show that DASC can inherit the strong controllability of weighted decoding, 
while also achieving a natural solution of multi-attribute control with the 
interpolation of multiple attribute embeddings in the space. Moreover, 
the shared attribute token embedding also alleviates over-parameterization, 
and improves the robustness of the model.

\begin{figure}[ht]
    \centering
    \includegraphics[width=0.75\columnwidth]{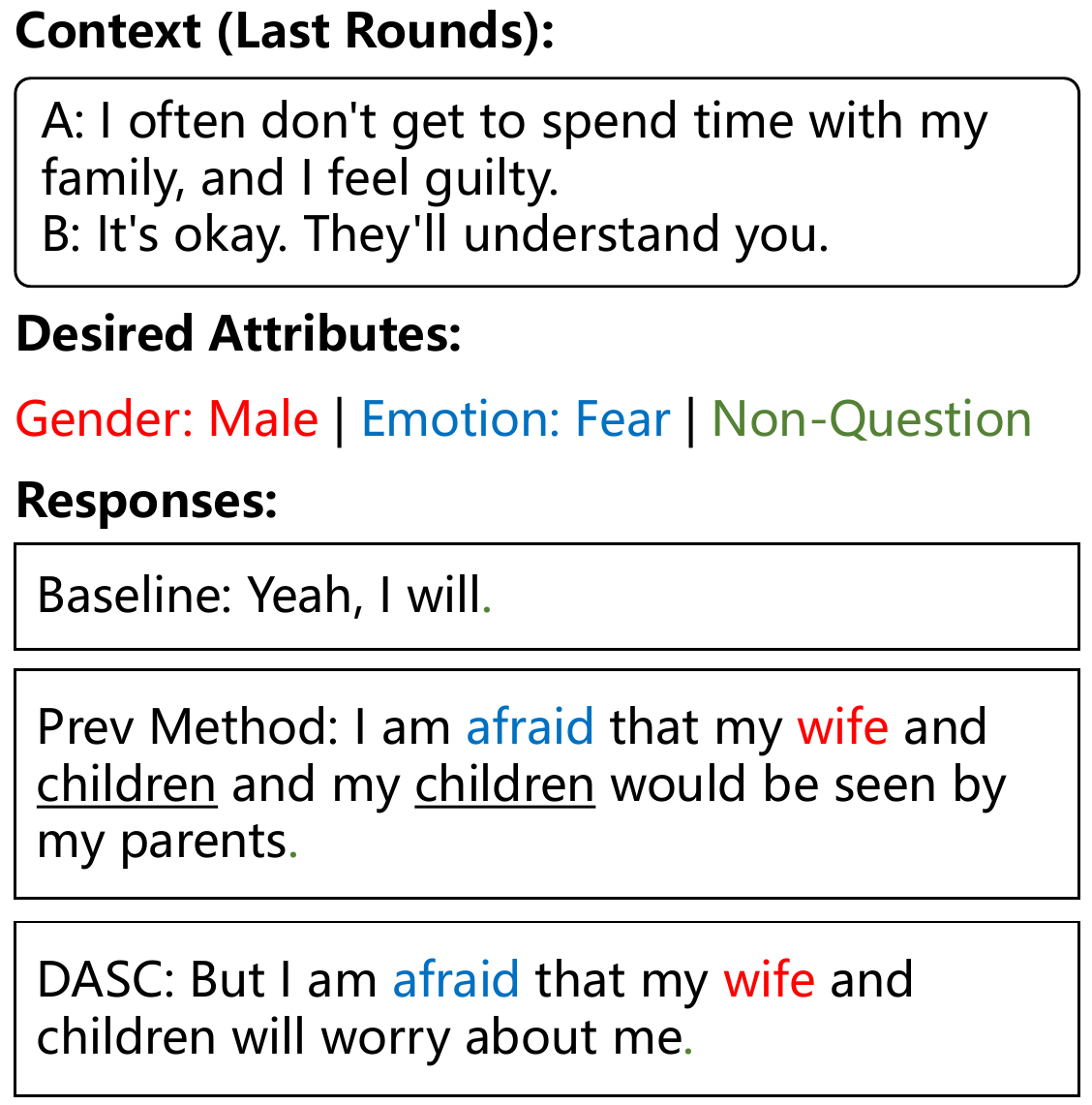}
    \caption{An example of multi-attribute controllable dialogue generation. 
The baseline system doesn't attempt any control and produced a dull response, 
while a previous method of attribute control generated a repetitive and illogical text. DASC successfully gives a response that is both fluent and 
correctly attributed.}
    \label{fig:teaser_example}
\end{figure}

We experiment on an attribute-rich open-domain dialogue dataset \citep{xu2022long} for the simultaneous control of 3 attribute aspects: Gender Style (male, female, neutral), Emotion (8 classes), and a simple division of Dialogue Act 
(question VS non-question). 
As exemplified in \figref{fig:teaser_example}, 
compared to previous methods, DASC achieves strong controllability while 
avoiding low-quality generations in the compositional controlling task. 
Visualization of the attribute token embeddings (in \figref{fig:token_emb}) 
exhibits specific patterns 
that benefit the controlling, compared to the general LM token embeddings. 
We further conducted a robustness test in a out-of-distribution setting and validated that DASC's controllability generalizes.
Our contributions are as follows: 1) We propose semantic space grounded weighted decoding for controllable dialogue generation, which can intuitively solve the multi-attribute control task with the interpolation of embeddings in the space; 2) DASC uses smaller number of parameters than other weighted decoding alternatives while achieving better performance with the design of shared attribute embeddings; 3) DASC can achieve high accuracy on the simultaneous control of 3 aspects while also preserving competitive generation quality in both conventional test settings and out-of-distribution robustness tests.

\section{Method}
\label{sec:method}
In this section, we will first define our task and weighted decoding method for controllable generation as background. Then we will introduce the proposed \textbf{DASC} framework. 
\subsection{Task Definition}

Given a dialogue \textbf{context} $C$ and \textbf{attributes} $A = (a_1, a_2, ..., a_K)$, \textit{Controllable Dialogue Generation} aims to generate a \textbf{response} $R = (r_1, r_2, ..., r_N)$ that is consistent with the context 
and carries the attributes.\footnote{In our work, we make a pre-assumption that attributes 
are provided by a dialogue policy, and do not include end-to-end scenarios.}
There can be multiple \textbf{aspects} grouping the attributes, where in this work we will focus on \textit{Gender Style, Emotion, and Dialogue Act}. An aspect covers multiple related attributes, such as \textit{happiness, sadness} for the Emotion aspect. Each attribute can take three values: 1 means to use the attribute, 0 means not to use and $\phi$ means the attribute is not
applicable to the response.

\subsection{Weighted Decoding for Controllable Generation}
\label{sec:weighted_decoding}

Standard, non-controllable, dialogue generation can be formulated with the standard conditional language modeling objective: $L_{CLM} = -\sum_{n=1}^N log P(r_n|r_{1:n-1}, C)$

We can use a transformer-based encoder-decoder architecture like BART \cite{lewis2020bart} to model this, where the encoder encodes the context into hidden states as condition for the decoder to generate the response. We will omit $C$ below for brevity.

In controllable dialogue generation, we additionally introduce attributes in the generation condition. Suppose we are generating with a single attribute $a$, then the objective is to model $P(r_n|r_{1:n-1}, a)$. Using Bayes' rule, this can be converted to:
\begin{equation}
    \small
    P(r_n|r_{1:n-1}, a) \propto P(r_n|r_{1:n-1}) P(a|r_{1:n-1}, r_n)^{\alpha}
    \label{eqn:single_wd}
\end{equation}
where $\alpha$ is a hyperparameter that can adjust the \textit{control strength}. This means that we can decompose the generation probability into the standard CLM probability weighted by the prediction of another token-wise attribute classifier during decoding. Methods established on such decomposition are thus called \textbf{Weighted Decoding} models.

Director \citep{arora2022director}, a representative weighted decoding method, implements the attribute classifier as a linear layer on top of the decoder hidden states. A binary classification is performed on determining whether the generated sentence reflects the desired attribute (e.g. happy or not) at each step. For tokens in the sentence from training set, they can be trained with the attribute of the whole sentence using Binary Cross Entropy (BCE) loss. We denote this token-level loss as $L_{t}$. 

\begin{equation}
    \begin{aligned}
        L_{t} &= BCE(P(a | r_{1:n-1}, r_n)) \\
                  &= BCE(\sigma([W_a h_n]_{r_n}))
    \end{aligned}
    \label{eqn:clf_t}
\end{equation}
where $h_n \in \mathbb{R}^{d}$ is the hidden state for the $n$-th token, $W_a \in \mathbb{R}^{|V| \times d}$ is the learnable weight matrix for attribute prediction given the generation of each token in the vocabulary, and $[\cdot]_{r_n}$ denotes the index selection with the next token $r_n$. 
Note that it only gathers the attribute logits with the token $r_n$ 
in the ground truth response. For the other $|V|-1$ tokens in the vocabulary 
$V$, they have no label and cannot get trained. Therefore, it uses an 
extra regularizer to train the prediction on these tokens to as close to 
0.5 as possible with MSE loss.  

When dealing with multi-attribute control, we can extend 
\eqnref{eqn:single_wd} by introducing the product of multiple attribute 
classifiers, assuming the conditional independence of attributes:

\begin{equation}
    \small
    P(r_n|r_{1:n-1}, a) \propto P(r_n|r_{1:n-1}) \prod_{\substack{k=1\\a_k\ne \phi}}^{K} P(a_k|r_{1:n})^{\alpha}
    \label{eqn:multi_wd}
\end{equation}

The product of probabilities is usually implemented with the 
summation of logits: 

\begin{equation}
    \small
    \delta(r_n|r_{1:n-1}, a) = \delta(r_n|r_{1:n-1}) + \alpha \sum_{\substack{k=1\\a_k\ne \phi}}^{K} \delta(a_k|r_{1:n})
    \label{eqn:multi_wd_logits}
\end{equation}

Existing works have implemented such an extension with either multiple forward passes through an 
attribute-conditioned language model \citep{lin2021plug} or one pass 
of multiple models \citep{liu2021dexperts}, which can all be very costly as the number of attributes grows. 
Here we introduce a relatively simple extension of Director, where we just add $K$ linear classifier heads to 
make the prediction of multiple attributes. We will refer to this 
simple extension as M-Director, or just Director for simplicity.
Note that though more efficient than previous methods, M-Director will still introduce $d \times |V| \times K$ extra parameters.
Given that $|V|$ is usually as large as tens of thousands, this model will 
have enormous number of parameters making it inefficient to train or infer, 
and also prone to overfitting. 

\subsection{Dialogue Attribute Space Controller}
\label{sec:dasc_method}
We hypothesize that the above typical methods of weighted decoding may 
not be the most efficient approach to learn the token-level attribute 
semantics, especially in multi-attribute cases. 
The learning objective is imposed on a single token in the target sentence, 
while all  other tokens are regularized equally. This is not usually reasonable, as some tokens similar to the target token should also have high probabilities 
given the attribute while other tokens different from it are less likely to be 
generated. For example, for the first token in a \textit{happy} response ``nice to meet you'', ``glad'' will also be a reasonable alternative, 
while ``sad'' is not, but their attribute label in the training will both 
be 0.5.

We can fix this counter-intuition in a high-dimensional space. 
On the one hand, each token has an $p$-dim embedding that encodes its attribute 
semantics (\textit{Attribute Token Embedding}, $ATEMB$). 
On the other hand, the hidden states from the LM ($h_n$) are also projected 
to the same space with attribute-specific linear layers 
($W^k \in \mathbb{R}^{p \times d}$) to get 
\textit{Attribute Context Embedding}, $\hat h^k_n = \hat W^k h_n$.
Thus different vectors in the space convey different semantics, 
and we call this space \textit{Attribute Semantic Space}. 

To leverage this latent space for weighted decoding, for each $\hat h^k_n$, we find its attribute-related tokens according to embedding similarity in the space, and assign them higher weights during decoding. Specifically, it is accomplished with a dot-product based token-level attribute classifier.

\begin{equation}
    \small
    \delta(a_k|r_{1:n}) = \hat h^k_n \cdot ATEMB(r_n)
    \label{eqn:matching_for_logits}
\end{equation}

In this case, when a token is trained with high probability for 
certain attribute, its neighbors in the attribute space will also 
have higher probabilities. This alleviates the limitation of previous 
weighted decoding methods, and eliminates the need for regularizers on 
other tokens. Further, when applying this to multi-attribute weighted decoding, 
we get: 

\begin{equation}
    \small
    \begin{aligned}
        \delta(r_n|r_{1:n-1}, a) &= \delta(r_n|r_{1:n-1}) \\
                                 &+ \alpha (\frac{1}{K} \sum_{\substack{k=1\\a_k\ne \phi}}^{K} \hat h^k_n) \cdot ATEMB(r_n)
    \end{aligned}
    \label{eqn:dasc_logits}
\end{equation}
where the parenthesized part in the second term can be interpreted as the 
average/equal-weight interpolation of multiple attribute context embeddings.
\footnote{It is possible to assign different weights for each embedding in interpolation, and we leave it for future works.}
This formulation suggests that if the attribute space is properly learned and represented, the embedding interpolation will precisely reflect the semantics of the desired attributes, and then DASC can realize reasonable attribute combinations. 

\begin{figure}[t]
    \centering
    \includegraphics[width=1.0\columnwidth]{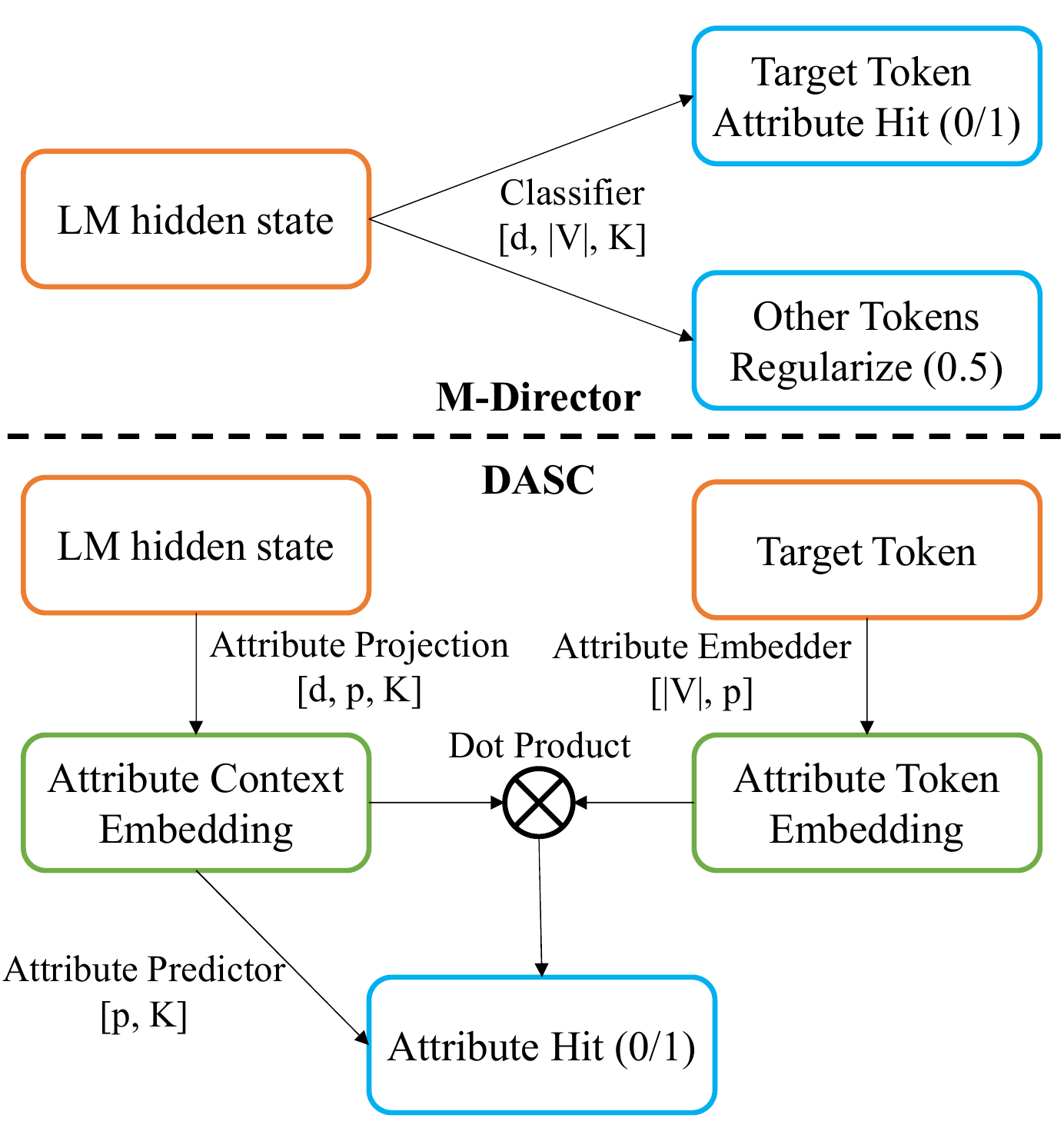}
    \caption{Framework comparison between M-Director and DASC. M-Director uses a classifier head to conduct binary attribute hit classification for each token in the target sentence, and impose regularization for other tokens. 
DASC projects both LM hidden state and the target token to the attribute space, and uses their dot product for the classification of attribute hit. 
For each parameterized model component, we show its shape in square brackets.}
    \label{fig:dasc_illustration}
\end{figure}

To assist the learning of attribute embeddings, we introduce another linear layer on top of the attribute context embedding at each step to directly predict the attributes of the complete response. This can help better align the attribute 
context embeddings with the corresponding region for its attributes. 
We denote the new the sentence-level classification loss as $L_{s}$. 
For clarity, we give its formulation in the single-attribute case, 
which can be simply extended to multi-attribute scenarios with the 
summation over all non-empty attributes.

\begin{equation}
    \begin{aligned}
        L_{s} &= BCE(P(a | r_{1:n-1})) \\
                  &= BCE(\sigma(v_a \cdot \hat h_n))
    \end{aligned}
\end{equation}
where $v_a \in \mathbb{R}^{p}$ is the learnable weight for attribute prediction. Compared with $L_{t}$ (\eqnref{eqn:clf_t}), it is a sentence-level classification task independent of $r_n$, which can also be interpreted as predicting the prior probability of the attribute before deciding the next token to generate, and thus the parameters do not scale with $|V|$. Then the final loss is: $L_{train} = L_{CLM} + \beta (L_{s} + L_{t})$, where $\beta$ is a hyperparameter that controls the weight of attribute-related losses.

We name the proposed framework as Dialogue Attribute Space Controller (\textbf{DASC}). The illustration of DASC and its comparison with M-Director is shown in Figure \ref{fig:dasc_illustration}. DASC introduce fewer parameters ($d\times p \times K + |V| \times p$)  than M-Director ($d \times |V| \times K$). Since we set $p << |V|$, the parameters of attributes projections will be much smaller. And when we deal with $K > 1$, the shared token embeddings across attributes will also save parameters, while the parameters of attribute predictor are almost negligible. 


\section{Experiments}
\label{sec:exps}
In this section, we will conduct experiments to examine the following hypotheses: (1) DASC can achieve strong controllability while also preserving good 
generation quality in multi-attribute controllable generation; 
(2) DASC's performance benefits from the meaningful representations in the 
attribute semantic space, and reduction in parameters; 
(3) DASC can also be flexibly extended for other control tasks like the 
composition of multiple emotions or adopting certain strategies for 
emotional support. 

\subsection{Experiment Settings}
We conduct experiments on the \textit{self} split of the DuLemon 
dataset~\citep{xu2022long}, which is a Chinese open-domain dialogue dataset 
that is rich in personalized content so that we can find the various 
attributes we would like to control. We split the data to train/dev/test set 
into 352,999, 2439, 2412 utterances each. Since the original dataset do not 
contain annotations of control attributes, we develop a few classifiers, 
one for each type of attributes, to label the dataset. 
For gender style (male, female, neutral), we use 
the dataset released by \citet{su2020stylistic} to train a MacBERT classifier \citep{cui2020revisiting}, 
which achieved accuracy=94.98\%. For emotion, we 
follow \citet{zhou2018emotional} and use the NLPCC2013 and NLPCC2014 dataset 
(8 emotion classes) to train another MacBERT classifier, which has an 
accuracy of 93.96\%. For the question dialogue act (question VS non-question), 
we simply use a heuristic for labeling: if the sentence contains a question 
mark(?) we will consider it a question and otherwise non-question. 
We then use these 3 classifiers to assign each response in the dataset 
with the 3 types of attributes (13 of them in total). 

\subsubsection{Competing Methods}
We compare the proposed DASC framework with representative methods from different types of controllable generation methods. We use the \texttt{fnlp/bart-base-chinese} \citep{shao2021cpt} model as the backbone for all competing methods~\footnote{We also conduct experiments with those leveraging persona description texts, including BoB \cite{song2021bob} and prompting with ChatGPT~\cite{openai2022:chatgpt}, although they may not be especially suitable for controlling the sparse attributes here. We will provide the experimental details and automatic evaluations in Appendix.}:
\textbf{Baseline} Simply fine-tuning the backbone on the dataset without utilizing the control attributes.
\textbf{Rerank} Using top-$p$ sampling \citep{holtzman2019curious} on the baseline model to produce 5 response candidates for each context, and attribute classifiers (here are the same separate models we've used for auto-annotations) to rerank the candidates. Following \citet{thoppilan2022lamda}, we use the sum of predicted probabilities in each aspect for ranking. 
\textbf{CTRL} We re-implemented \citet{keskar2019ctrl}'s method for dialogue generation by defining 3 groups of special control codes for each aspect, and appending the corresponding 3 attribute tokens to each dialogue context during fine-tuning.
\textbf{Director} The multi-attribute extension of Director \citep{arora2022director} discussed in \secref{sec:weighted_decoding}. We provide more experimental details in Appendix.

\subsubsection{Evaluation}
\paragraph{Automatic Evaluation} To evaluate the controllability, we use the same attribute classifiers as those used for labeling the dataset to calculate the accuracy of attributes in the generation (Acc$_G$, Acc$_E$, Acc$_Q$ for gender, emotion and question, respectively). For the generation quality, we use BertScore (BScore) \cite{zhang2019bertscore} to evaluate generation's similarity to reference response, and Distinct-2 \cite{li2016diversity} for diversity. 

\paragraph{Human Judgement} We sampled 100 contexts from the test set for human evaluation. Since the distribution of the original test set is extremely skewed, we've specified a constraint for more balanced distribution over all emotions during sampling, so as to ensure the representativeness of the evaluation (21 none, 16 sadness, 16 disgust, 16 happiness, 16 like, 5 anger, 5 surprise, 5 fear). We invited 2 volunteers who are native Chinese speakers to evaluate 
each generation from 3 perspectives. 
\textbf{Attribute Accuracy}: if the response conveys the given attribute. 
\textbf{Sensibleness}$_{(1-4)}$: if the response is fluent, coherent with the 
context, and accords with commonsense. 
\textbf{Interestingness}$_{(1-4)}$: whether the response is specific, novel 
and can encourage more interesting conversation continuation. 

\subsection{Results}
\label{sec:results}
\begin{table}[]
    \small
    \centering
    \begin{tabular}{rccccc}
    \hline
             & BScore      & Dist-2         & Acc$_G$         & Acc$_E$         & Acc$_Q$          \\ \hline
    Baseline & 68.18          & 19.25          & 68.49          & 46.31          & 69.61           \\
    Rerank   & 69.23          & 19.28          & 75.46          & 54.93          & 82.42           \\
    CTRL     & \textbf{71.09} & 18.91          & 85.32          & 77.49          & \textbf{100.00} \\
    Director & 69.54          & {\ul 21.40}    & {\ul 95.81}    & \textbf{86.73} & \textbf{100.00} \\
    DASC     & {\ul 70.42}    & \textbf{21.94} & \textbf{95.85} & {\ul 86.07}    & \textbf{100.00} \\ \hline
    \end{tabular}
    \caption{Automatic evaluation results on DuLemon test set. The best results are in bold, while the second results are underlined.}
    \label{tab:auto_results}
\end{table}

Automatic evaluation results are shown in Table \ref{tab:auto_results}. We can see that \textit{Rerank} failed to show strong controllability because the base model struggles to produce attributed ranking candidates without finetuning with the attributes. \textit{CTRL} leveraged the attributes in finetuning, and achieved better control accuracy and BertScore, but it doesn't produce more diverse responses overall. Both \textit{Director} and \textit{DASC} exhibit the best controllability, and \textit{DASC} produces more diverse and reasonable responses according to Distinct-2 and BertScore. 

\begin{table}[th]
    \small
    \centering
    \begin{tabular}{rccccc}
    \hline
                & Acc$_G$       & Acc$_E$       & Acc$_Q$  & Interest      & Sensible      \\ \hline
    Baseline & 0.80          & 0.55          & 0.64          & 2.04          & \textbf{3.46} \\
    Rerank   & 0.81          & 0.62          & 0.82          & 2.13          & 3.44          \\
    CTRL     & 0.85          & 0.82          & \textbf{0.97} & 2.24          & \textbf{3.46} \\
    Director & 0.87          & 0.87          & 0.96          & 2.25 & 3.26 \\
    DASC     & \textbf{0.88} & \textbf{0.88} & \textbf{0.97} & \textbf{2.37} & 3.28 \\ \hline
    \end{tabular}
    \caption{Human Judgement on DuLemon test set.}
    \label{tab:human_results}
\end{table}

We then show human judgement results in Table \ref{tab:human_results}. The inter-annotator agreement for Acc$_G$, Acc$_E$ and Acc$_Q$ are 0.65, 0.55 and 0.64 in Cohen's $\kappa$, which indicates moderate to substantial agreement. The agreement of $Interestingness$ and $Sensibleness$ is 0.48 and 0.44 in Pearson's $r$. This is hardly
surprising because the latter two perspectives are highly subjective. 
The evaluation on attribute accuracies is similar to the automatic results, 
except that the accuracy of gender drops slightly. We find that human evaluators can spot errors related to gender stereotype \citep{bolukbasi2016man}, like generating soldier for male style and baby-carer for female, where these occupations should be gender-neutral. The annotators also check questions 
without a question mark, which explains the slight difference in Acc$_Q$.

Overall, the rankings of controllability still hold according to human evaluation, with DASC performing the best. Baseline, Rerank and CTRL have slightly better \textit{Sensibleness} than weighted decoding methods, which agrees with the commonly observed controllability-quality trade-off in previous literature \citep{dathathri2019plug,yang2021fudge,qian2022controllable}. All controllable generation methods achieved higher \textit{Interestingness} score than baseline, which supports the benefits of controllable generation. DASC achieved the best \textit{Interestingness} given similar attributes accuracy as Director, indicating the effectiveness of attribute semantic space, which can establish better representations of attribute semantics and a more reasonable approach to compose the control attributes in weighted decoding. 


\subsection{Robustness Test}
In previous experiments, the control attributes provided to the model come from the reference response. Therefore, models may coincidentally hit the desired attributes when generating the most likely response to the context, without truly reliable controllability for arbitrary given attributes. Hence, we further conduct experiments to test the robustness of the controllable generation methods in out-of-distribution scenarios. 

Specifically, we sampled 100 contexts from the test set, and give the models each of the 8 emotions as the generation condition, paired with the original gender and question act\footnote{We do not change these 2 attributes as they are sometimes determined given the context.}. We then use greedy decoding to generate response for each (context, attributes) pair and conduct similar automatic and human evaluation on the 800 generations. 

\begin{table}[th]
    \small
    \centering
    \begin{tabular}{r|cc|cc}
    \hline
             & Dist-2         & Acc$_E$            & Interest      & Sensible      \\ \hline
    Rerank   & 17.55          & 17.00          & -             & -             \\
    CTRL     & 21.07          & 43.38          & 1.91          & \textbf{3.00} \\
    Director & \textit{34.73} & 61.88          & 1.62          & 2.27             \\
    DASC     & \textbf{26.71} & \textbf{65.38} & \textbf{2.08} & 2.82 \\ \hline
    \end{tabular}
    \caption{Robustness test results.}
    \label{tab:robustness}
\end{table}

\begin{figure*}[t]
    \centering
    \begin{subfigure}{}
        \centering
        \includegraphics[width=.48\linewidth]{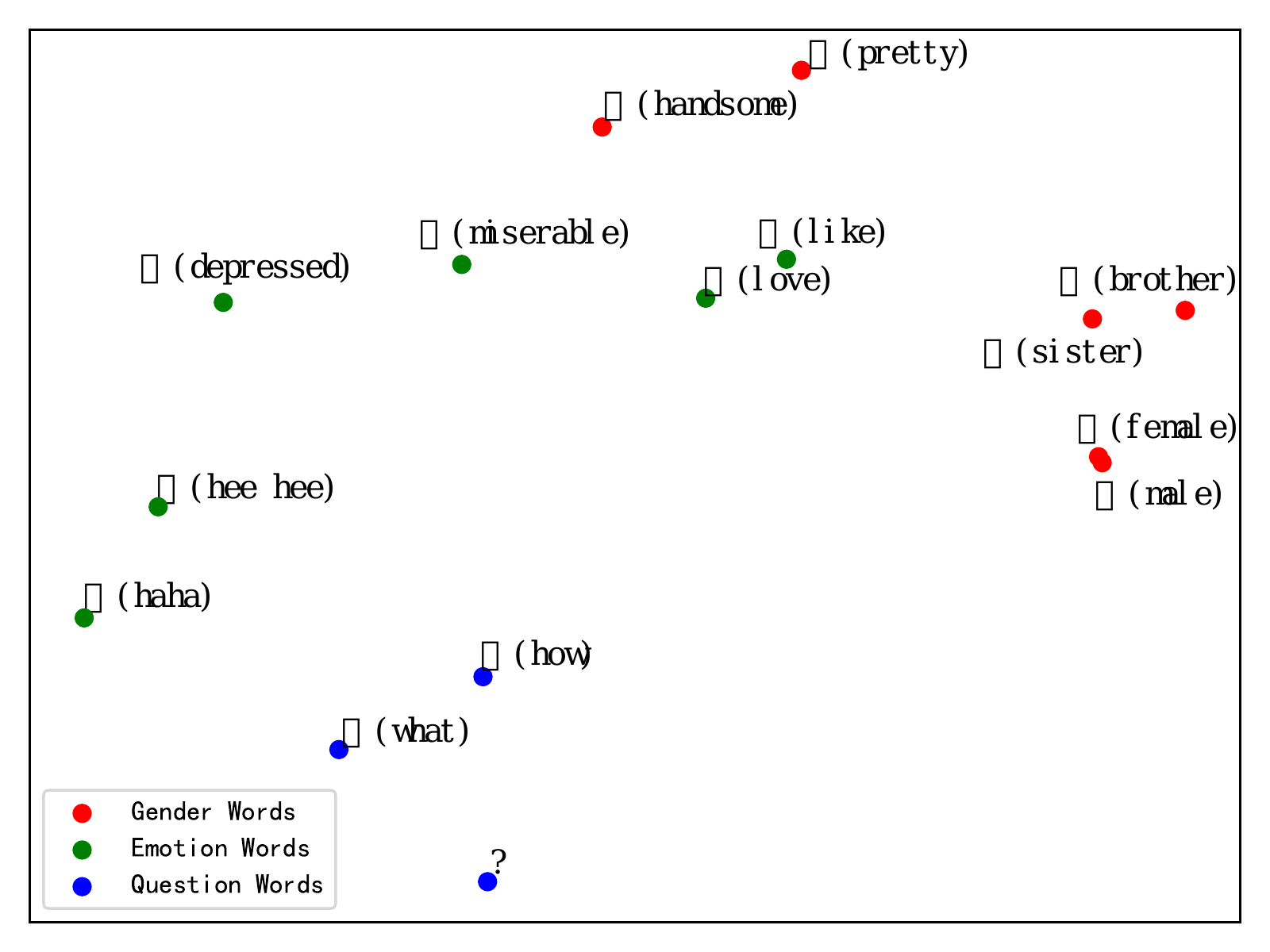}  
    \end{subfigure}
    \begin{subfigure}{}
        \centering
        \includegraphics[width=.48\linewidth]{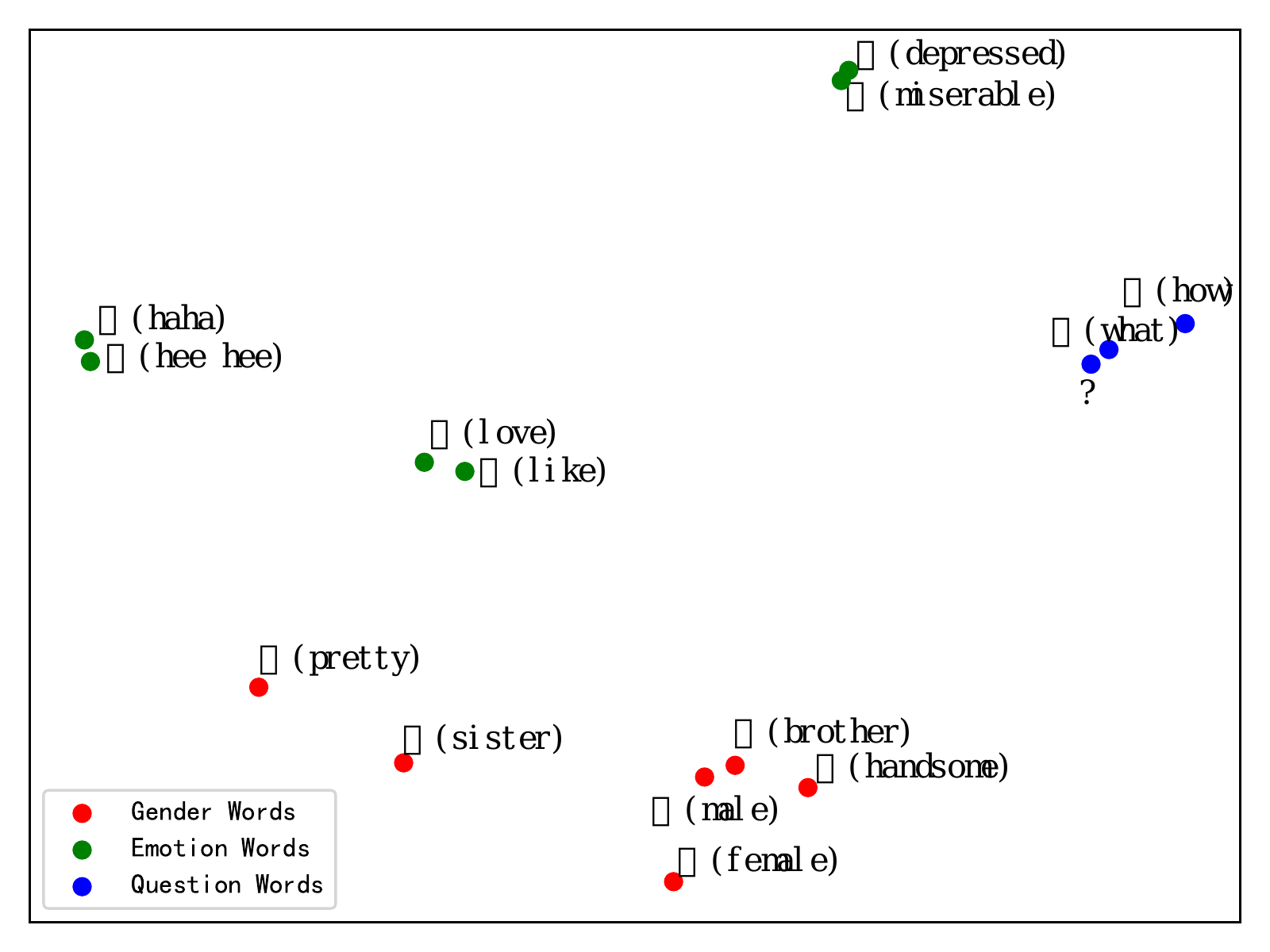}  
    \end{subfigure}
    \caption{Comparison of two sets of token embeddings with t-SNE visualization: those from the language model (left) and from the attribute semantic space (right).}
    \label{fig:token_emb}
\end{figure*}

\begin{figure}[t]
    \centering
    \includegraphics[width=0.9\columnwidth]{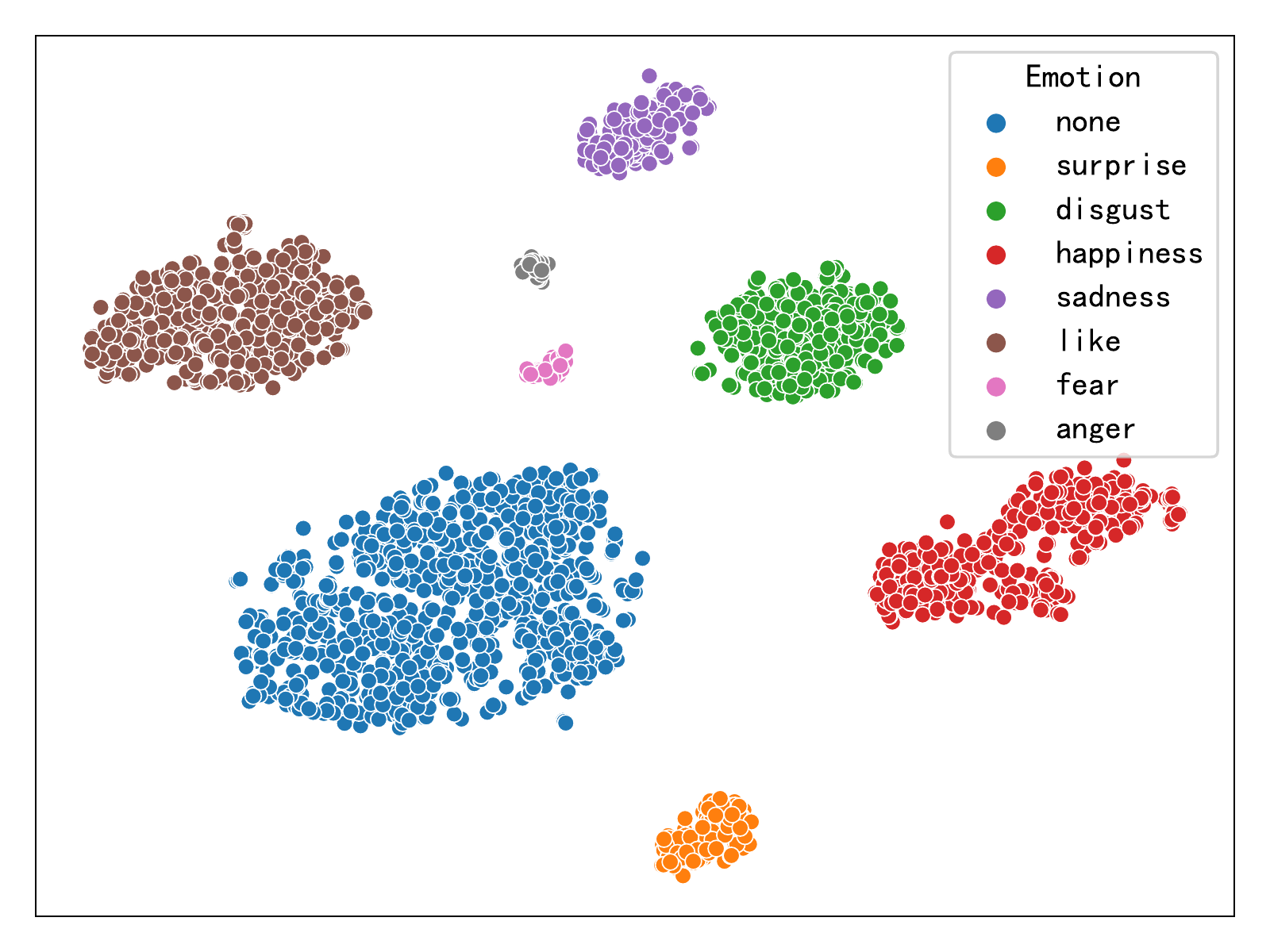}
    \caption{The t-SNE visualization of attribute context embedding of responses with different emotions.}
    \label{fig:emotion_context_emb}
\end{figure}

Table \ref{tab:robustness} shows the robustness test results.\footnote{BertScore is not reported here, as the model can be directed towards attributes different from the ground truth, invalidating the similarity-based metric as a proxy for generation quality.} Compared with Table \ref{tab:auto_results}, we can see that the emotion accuracy of Rerank and CTRL dropped significantly, which shows that their controllability is not generalizable. Another notable phenomenon is the abnormal \textit{Distinct-2} achieved by Director. We then further analyze their performance with human evaluation (excluding Rerank as it fails to control attributes). We found that Director frequently generate ungrammatical, illogical and repetitive long responses (like the second response in Figure \ref{fig:teaser_example}). Director's loss in emotion accuracy is also higher than DASC, indicating that it may overfit the training distribution given its large parameters, and thus performs worse in this out-of-distribution setting. Compared to CTRL, DASC has lower \textit{Sensibleness} but higher \textit{Interestingness}, when it also has a significant advantage in diversity and controllability.

\subsection{Semantic Space Visualization}

For a clear understanding of how the proposed attribute semantic space can 
help controllable generation, we visualize them in 2D space with 
t-SNE~\citep{van2008visualizing}. First, we visualize the attribute token 
embeddings of some representative attribute-related tokens, and also compare 
them with the corresponding embedding in the original LM (\figref{fig:token_emb}). Comparing the two figures, we can see that (1) The token embeddings from different 
aspects are more separable in the attribute space 
(see points with different colors), while tokens in the same aspect
are closer despite the difference in other linguist features like 
part-of-speech (like `handsome' and `male'). 
(2) The token embeddings from different attributes of the same aspect
are also distinguished in the attribute space (like `male'-`female', `love'-`miserable'). These characteristics enable DASC to successfully control the generation of distinctive attributes and compose attributes from different aspects.

Next, we also visualize the attribute context embedding. Specifically, we take the responses with certain attribute in the dev set of the dataset, feed them into the model and average attribute context embeddings at each decoder token as sentence-level representations, and pair them with the sentence-level attribute annotations for analysis. For brevity, we only show the visualization with emotion labels in Figure \ref{fig:emotion_context_emb}, and provide those with gender and question act labels in Appendix. We can see that the context embeddings from sentences with different emotions are clearly separated in the space, which supports the strong controllability of DASC with multiple attributes.

\subsection{Parameter Analysis}
\label{sec:parameter_analysis}
As analyzed before, DASC can use a relatively smaller amount 
of parameters to implement weighted decoding for multi-attribute controllable 
generation. Here we study the effect of number of parameters by adjusting the 
dimension of the attribute space $p$, and comparing with baseline and 
M-Director which uses no/large amount of parameters for attribute control. We use BertScore to evaluate the generation quality and 
the average control accuracy on 3 aspects to reflect controllability.

\begin{table}[th]
    \small
    \centering
    \begin{tabular}{lccc}
    \hline
    Method      & \#params        & BScore         & Avg Acc        \\ \hline
    baseline    & -               & 68.18          & 61.47          \\
    DASC ($p$=512) & 15.94M          & 70.18          & 92.56          \\
    DASC ($p$=1024)& 31.88M          & 70.12          & 92.72          \\
    DASC ($p$=2048) & 63.75M          & \textbf{70.42} & 93.97          \\
    DASC ($p$=4096)& 127.50M         & 70.26          & \textbf{94.42} \\
    Director    & 210.98M         & 69.54          & 94.14          \\ \hline
    \end{tabular}
    \caption{Effect of the number of extra parameters for controllability and generation quality.}
    \label{tab:num_params}
\end{table}

Results are shown in Table \ref{tab:num_params}. Comparing DASC with different $p$, we can see that larger amount of parameters can generally improve the model's controllability, but even a relatively small $p$ ($p$=512) is already capable to achieve high control accuracy. For generation quality, Director, which additionally uses nearly twice the parameters of the base model (210.98M vs. 116.26M), may have been over-parameterized and thus harms its generation quality. A moderate size of DASC can achieve the best BertScore, but smaller ones do not significantly degrade the performance. This suggests tha DASC can be a promising candidate in application, given its parameters are fewer than alternatives and are orders of magnitude fewer than LLMs that has generally over 6B parameters.

\subsection{Case Study}

\begin{figure}[h]
    \centering
    \includegraphics[width=0.75\columnwidth]{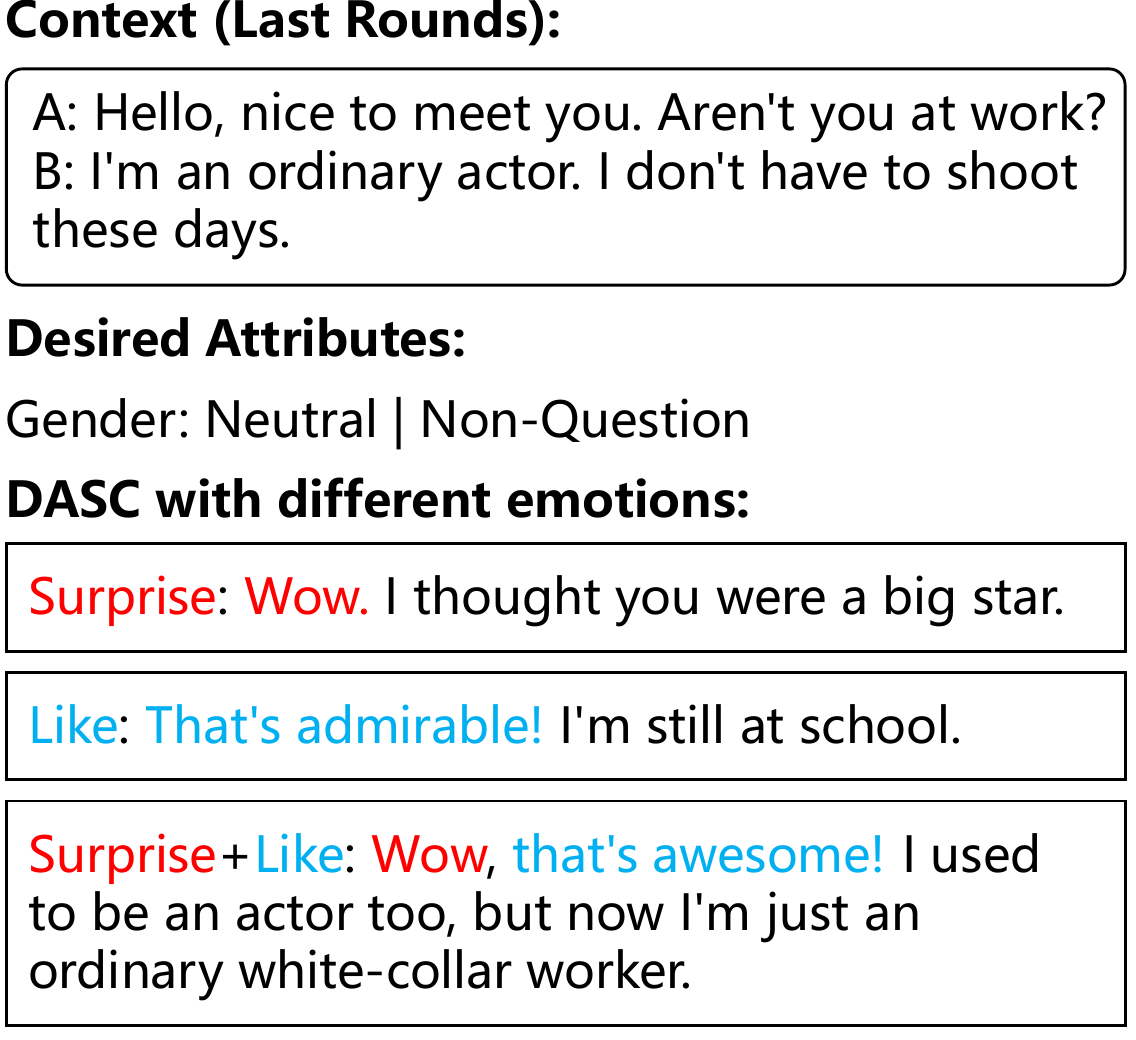}
    \caption{DASC generates different responses to the same context given different emotions and their composition as control attributes.}
    \label{fig:compose_example1_en}
\end{figure}

Besides multi-aspect control as shown in Figure \ref{fig:teaser_example}, 
we also show a proof-of-concept application that DASC can naturally blend two emotions in one generated response. We can simply achieve this by setting both attributes' value as 1 instead of $\phi$. The results are shown in Figure \ref{fig:compose_example1_en} and Figure \ref{fig:compose_example2}. We can see that DASC can successfully generate responses with either single emotion or the combination of both emotions, where the later can produce potentially more vivid response.

\subsection{ESConv Experiment}

To further explore the potential of DASC, we also experimented on another dataset ESConv \citep{liu2021towards}. It is an English dataset that aims to provide emotional supports to help seekers with 8 defined strategies. Here we use the human annotated strategy labels as the control attributes, and experimented with 3 methods: \textbf{Baseline}, \textbf{CTRL} and \textbf{DASC}. We excluded Director here for its inefficiency. 
We report the automatic metric \textbf{Distinct-2} and human evaluated \textbf{Strategy Accuracy}, \textbf{Usefulness}$_{(1-4)}$ and \textbf{Sensibleness}$_{(1-4)}$. In Table \ref{tab:esconv_results}, we can see that the control of relatively complex strategies is harder, and thus the accuracy is lower than the previous experiment (Table \ref{tab:human_results}). Nevertheless, DASC still achieves reasonable control accuracy and outperforms other methods on all metrics. These results suggest that DASC is language-agnostic and can be effectively applied to many kinds of attribute controls. We provide more details and generation examples in Appendix.

\begin{table}[th]
    \centering
    \small
    \begin{tabular}{rcccc}
        \hline
             & Dist-2         & Acc           & Useful        & Sensible      \\ \hline
    Baseline & 19.28          & 0.27          & 1.92          & 3.30          \\
    CTRL     & 21.20          & 0.52          & 2.04          & 3.31          \\
    DASC     & \textbf{25.86} & \textbf{0.70} & \textbf{2.24} & \textbf{3.48} \\ \hline
    \end{tabular}
    \caption{Test results on ESConv.}
    \label{tab:esconv_results}
\end{table}

\section{Related Work}

Controllable generation has gained wide research interest recently. PPLM \citep{dathathri2019plug} proposed a plug-and-play framework to control the generation with an extra attribute classifier. Later research progress can be roughly divided into 3 categories. \textit{Reranking} methods leverage attribute classifiers to either simply rank the full generation candidates \citep{thoppilan2022lamda}, or partial generations for the guidance of future outputs \citep{yang2021fudge}. \textit{Integrated} methods integrate attribute-related trainable parameters into the generation model for fine-tuning, such as discrete control codes \citep{keskar2019ctrl} or continuous prompt prefix \citep{qian2022controllable}. \textit{Weighted Decoding} methods leverage token-level attribute classifiers to guide each decoding step. For example, \citet{krause2021gedi} and \citet{liu2021dexperts} utilized one/two additional class conditional language models to provide the attribute discrimination. Director \citep{arora2022director} integrates the attribute classifier as simple linear layers on top of LM hidden states.  

Multi-attribute controllable generation is relatively under-explored now. \citet{lin2021plug} proposed to extend weighted decoding for the multi-attribute case with the simple product of multiple attribute conditional language models. \citet{gu2022distributional} proposed a VAE-based method combined with an intersection-searching algorithm for multi-aspect controllable generation, but their method cannot simply apply to conditional generation tasks like dialogue generation. \citet{mireshghallah2022mix} proposed an energy-based controllable generation method that can combines multiple controls, but are mainly suitable for fixed-length generation. 

Controllable generation techniques are especially important in dialogue systems and the applications of several controlling aspects have been studied. For example, we may condition the generation with dialogue acts for the genuine reflection of the desired behavior \citep{wen2015semantically}, add emotions in the response to enhance the expressiveness of the bot \citep{zhou2018emotional}, and also impose personal profiles like gender \citep{su2020stylistic} and persona \citep{zhang2018personalizing} to establish a human-like companion.
Recent advance in LLMs has pushed the frontiers of dialog generation, 
enabling applications like role-playing with complex personality and 
memory~\citep{park2023generative}. However, their exorbitant
cost and privacy concerns make them less relevant in certain deployment 
scenarios.

Another line of controllable generation utilizes dense persona descriptions \citep{zhang2018personalizing}. This paradigm is capable of expressing rich persona information in free text, such as personal status, hobbies and occupations. The natural language form allows for integration of other language resources for an enhanced generation quality. For example, \citet{song2021bob} disentangles the task of persona consistency learning and response generation, and leverages non-dialogue NLI datasets to help the former and consequently enhance the latter. However, although attributes can also be expressed in free-text descriptions, they can contain noise, and making them less effective than attribute-specific methods, as shown in our experiments (Appendix \ref{sec:desc_control}). It would be promising to further combine the two paradigms for more general controllable generation \citep{tang2023enhancing}.
\section{Conclusion}

In this paper, we propose DASC, a novel framework for multi-attribute controllable dialogue generation. It is established on the weighted decoding paradigm for strong controllability and further grounds it in an attribute semantic space, which enables the simultaneous control of multiple attributes with the interpolation of multiple attribute embeddings. Experiments show that DASC can achieve strong controllability for multi-attribute generation while also preserving high quality in out-of-distribution scenarios. DASC is highly efficient given its much fewer number of parameters than alternatives and LLMs, making it an promising choice for deployment.

\section*{Limitations}

Some limitations of the proposed methods remain to be addressed 
in future research.

First, our experiment settings assume that the desired attributes are available for generation, which would require a separate dialogue policy to decide the 
attribute label provided to the model. Therefore, our model cannot be 
directly applied to end-to-end dialogue models, and may also be affected by 
the potential error propagation from the dialogue policy model. Since the intended use of DASC is to serve as a component of pipeline-style dialogue systems, these common issues in such systems are out of the scope of this work. 

Moreover, we require annotated datasets with multiple attributes for evaluation, which are rare. Therefore, we evaluate the capability of multi-attribute control mostly on one dataset. Experiments on more datasets, especially those with additional attributes may be required to further validated the efficacy of the proposed methods.

Last but not least, DASC is not directly applicable for controllable generation with free text as control signal, such as persona descriptions \cite{zhang2018personalizing}, which might limit its application range, though we may simply combine DASC with other techniques like concatenating the descriptions to achieve this goal, which will require further explorations. 

\section*{Ethics Statement}
The proposed method is utilized for the control of gender style. As we've noticed and discussed in Sec. \ref{sec:results}, the model may resort to gender stereotypes for generating responses in that gender. The potential reason is that the dataset used to train the classifier already contains gender-biased labels, and such biases are exploited by the classifier, and passed to the generation model through the automatic annotated labels. To avoid such effects, we may carefully clean the dataset for such biased labels \cite{gehman2020realtoxicityprompts}, or mine such biased tokens and penalize them during weighted decoding. We may also apply RLHF to further mitigate the biases \citep{ouyang2022training}.

Though the proposed method is mainly intended for improving the interestingness of the chatbot, and endowing the model with abilities like emotional support, such method may also be applied for vicious application. For example, they may introduce toxicity as an attribute and encourage the model to generate more toxic responses. Therefore, the application range of such techniques should be carefully restricted. 

We adhere to the license of the used datasets.


\bibliography{custom}
\bibliographystyle{acl_natbib}

\appendix

\section{Effect of Control Strength}

\begin{figure}[ht]
    \centering
    \includegraphics[width=1.0\columnwidth]{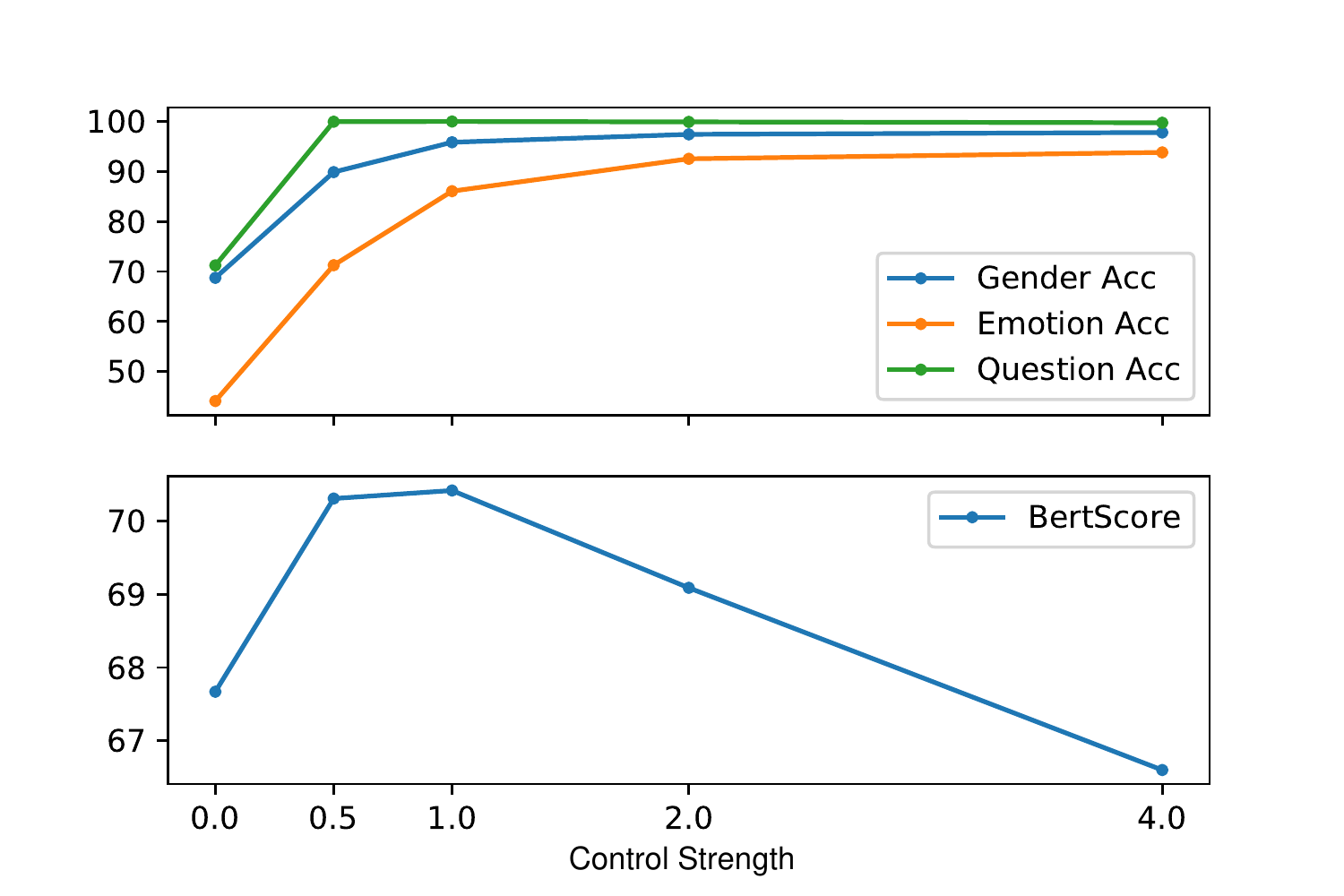}
    \caption{Effect of control strength on controllability and generation quality}
    \label{fig:parameter_tune}
\end{figure}

We show the effect of control strength $\alpha$ (\eqnref{eqn:dasc_logits}) on DASC's controllability and generation quality in Figure \ref{fig:parameter_tune}. From the trend shown in this figure, we can see that for \textit{Question} which is easy to control, we can already achieve perfect control with a low $\alpha$, while harder attributes like \textit{Emotion} would require a higher $\alpha$ to get a high success rate. Therefore, we may further hypothesize that a better balancing of the control accuracy of each attribute and the generation quality can be achieved by setting different control strengths for each aspect, like higher $\alpha$ for Emotion and lower $\alpha$ for Question. Careful tuning of the parameters or specific searching algorithms \citep{gu2022distributional} may serve the goal, and we leave this for future work.\footnote{In practice, we further multiply $\alpha$ by the number of attributes $K$ to adapt to the variable attribute numbers, which is not counted in \eqnref{eqn:dasc_logits} and Figure \ref{fig:parameter_tune} for clarity.}

\section{Experiment Details}

For all experiment methods, they use \texttt{bart-base}\footnote{\url{https://huggingface.co/facebook/bart-base}} as the base model. All models are fine-tuned on the dataset for 6 epochs. When conducting multi-aspect control for Director and DASC under weighted-decoding paradigm (\eqnref{eqn:multi_wd_logits}), we set the variables for the desired attribute as 1, and other variables as $\phi$. The decoding method is top-$p$ sampling with $p=0.5$. DASC uses control weight $\alpha=1$ and classifier loss weight $\beta=0.1$, similar as the previous experiments. All experiments of the paper are conducted on a Linux server, and each experiment is run on a single NVIDIA A100 GPU. To avoid overfitting, we select the checkpoint with the best BertScore on dev set for final testing. We fix the random seed in experiment and report the results coming from a single run. Below we provide the specific details for the experiments on ESConv.

For experiments on ESConv \citep{liu2021towards}, we use the latest released version\footnote{\url{https://github.com/thu-coai/Emotional-Support-Conversation}}, which has 1,300 conversations, and we split them into 1,100/100/100 train/dev/test set, which contains 15,605/1,403/1,369 utterances each. In human evaluation, we further sample 15 utterances for each of the 7 emotional support strategies defined in the dataset (except the vague \textit{Other} class), and get 105 utterances in total. The meaning of \textbf{Sensibleness}$_{(1-4)}$ is similar to the experiment in the previous dataset: if the response is fluent, coherent with the context, and accords with commonsense. By \textbf{Usefulness}$_{(1-4)}$, we consider if the response dives deep in the problem faced by the support seeker, is comforting, contains useful suggestions or encourages in-depth further discussions. 

\section{Experiments with Description Control}
\label{sec:desc_control}

Dense persona descriptions are another common form of control signal in dialog generation \citep{zhang2018personalizing}, and we can also convert the sparse attributes into descriptive texts to be compatible with these methods. Therefore, we now supplement new experiment results with two representative methods that leverages persona descriptions for control.

\paragraph{BoB} \citep{song2021bob} disentangles the task of persona consistency learning and response generation, and leverages non-dialogue NLI datasets to help boost the performance of consistency learning and finally improves personalized generation. 

To apply BoB on the dataset we've experimented with, we convert the discrete attribute annotations into textual descriptions with rules. For example, the male/female gender will be converted to "I'm a girl/boy.", a question will have the description "I want to ask a question". And for emotion, we fill them in the template "I feel \{emotion\}." We concatenate these 3 description texts as the persona text to be used by BoB. As is suggested in the official GitHub repository, we leverage the Chinese NLI dataset CMNLI \citep{xu-etal-2020-clue} as the auxiliary inference datasets.

\paragraph{ChatGPT} \citep{openai2022:chatgpt} is a representative Large-Language-Model (LLM) that can follow human instruction and give responses. Therefore, we can encode the attribute values into the ChatGPT system message to achieve control on them. We use the template:

\begin{quote}
    A dialog history is given below. Please act as the \{current speaker\} and respond to the next sentence with a \{dialog act\} in the voice of \{gender and emotion\}.
\end{quote}

Then we use dialog history as the user message, and let ChatGPT (gpt-3.5-turbo) give a following sentence. We use temperature=0 and stop=``\textbackslash n'', that is, greedy search for one line of text similar to the setting of the original dataset.

\begin{table}[h]
    \small
    \centering
    \begin{tabular}{cccccc}
    \hline
             & BScore         & Dist-2         & Acc$_G$         & Acc$_E$         & Acc$_Q$          \\ \hline
    Baseline & 68.18          & 19.25          & 68.49          & 46.31          & 69.61           \\
    CTRL     & \textbf{71.09} & 18.91          & 85.32          & 77.49          & \textbf{100.00} \\
    DASC     & 70.42          & 21.94          & \textbf{95.85} & \textbf{86.07} & \textbf{100.00} \\ \hline
    BoB      & 65.47          & 23.44          & 74.59          & 64.76          & 98.51           \\
    ChatGPT  & 66.21          & \textbf{30.98} & 69.49          & 56.88          & 98.22           \\ \hline
    \end{tabular}
    \caption{Automatic evaluation on the DuLemon test set with persona description-based controlling methods.}
    \label{tab:desc_control}
\end{table}

\begin{table}[h]
    \small
    \centering
    \begin{tabular}{ccc}
    \hline
            & Dist-2         & Acc$_E$         \\ \hline
    CTRL    & 21.07          & 43.38          \\
    DASC    & 26.71          & \textbf{65.38} \\ \hline
    BoB     & 24.25          & 30.13          \\
    ChatGPT & \textbf{37.02} & 30.00          \\ \hline
    \end{tabular}
    \caption{Automatic evaluation on the DuLemon robustness test with persona description-based controlling methods.}
    \label{tab:desc_control_robust}
\end{table}

Firstly, we can see that ChatGPT can generate highly diverse texts (with dist-2 on test set similar as the human ground truth, 30.98 VS 30.63). However, ChatGPT cannot achieve comparable controllability with finetuned methods like CTRL and DASC, especially on attributes whose manifestation cannot be easily described in the prompt (e.g. Gender and Emotion). 

Moreover, we can see that BoB also shows strong controllability compared to baseline and zero-shot ChatGPT. It also has higher generation diversity than other methods except ChatGPT, potentially due to the introduction of the auxiliary inference dataset. However, its controllability is relatively worse than DASC. Its generation quality is also poor, both reflected in the low BertScore and our manual check, where we find many influent cases like ``Yes. After all, I'm a lawyer, otherwise it would be hard to find a body.''. BoB also shows less sensitivity to the change of control signals, as is shown in the lower dist-2 and emotion accuracy in the robustness test. 

To conclude, we believe it is possible for description-based controllable generation methods like ChatGPT and BoB to perform better in the control of discrete attributes, but it would require significant efforts on prompt engineering (e.g. describe in more detail for ChatGPT, or make the description more similar to the auxiliary inference dataset for BoB). However, when we have relatively sufficient labeled data the discrete control attributes, DASC on `small' LMs will certainly be a simple and competitive choice.

\section{Examples and Visualizations}

In this section, we provide supplementary examples and figure visualizations. 

\begin{figure}[ht]
    \centering
    \includegraphics[width=1.0\columnwidth]{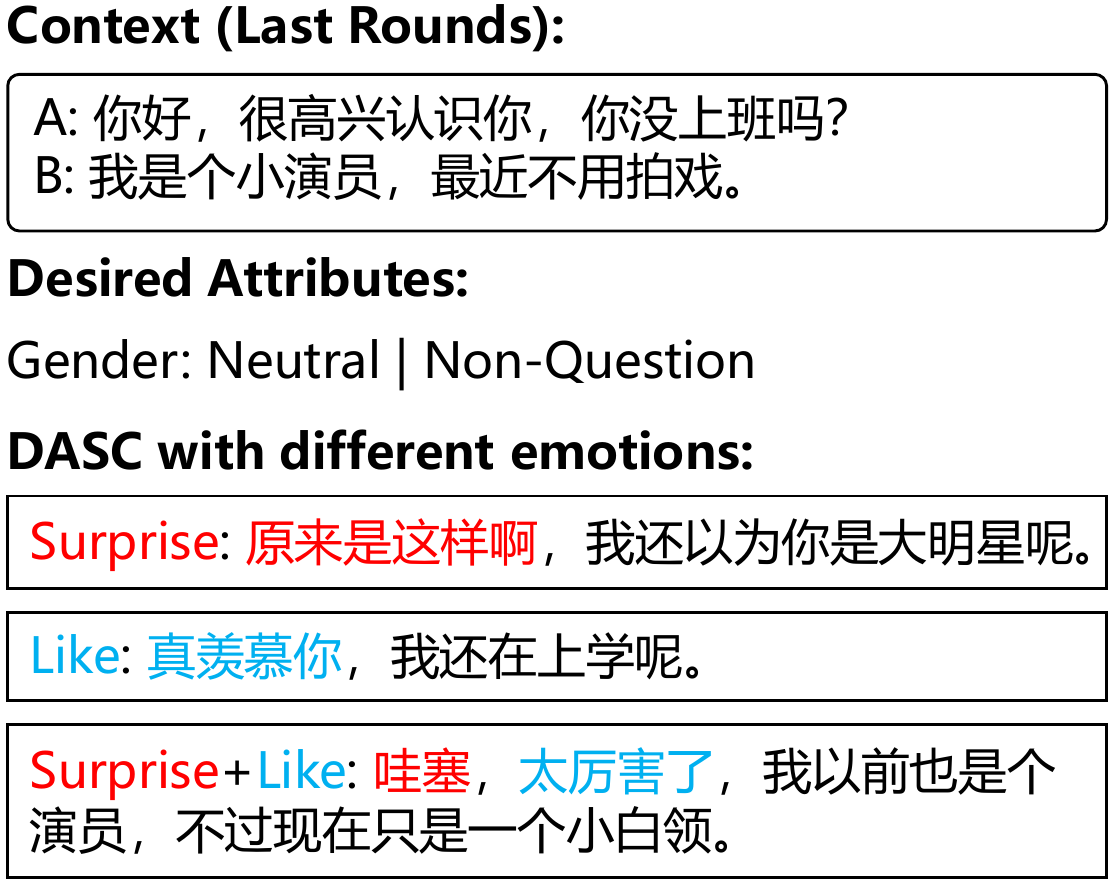}
    \caption{Original Chinese text for Figure \ref{fig:compose_example1_en}}
    \label{fig:compose_example1_zh}
\end{figure}

\begin{figure}[ht]
    \centering
    \includegraphics[width=1.0\columnwidth]{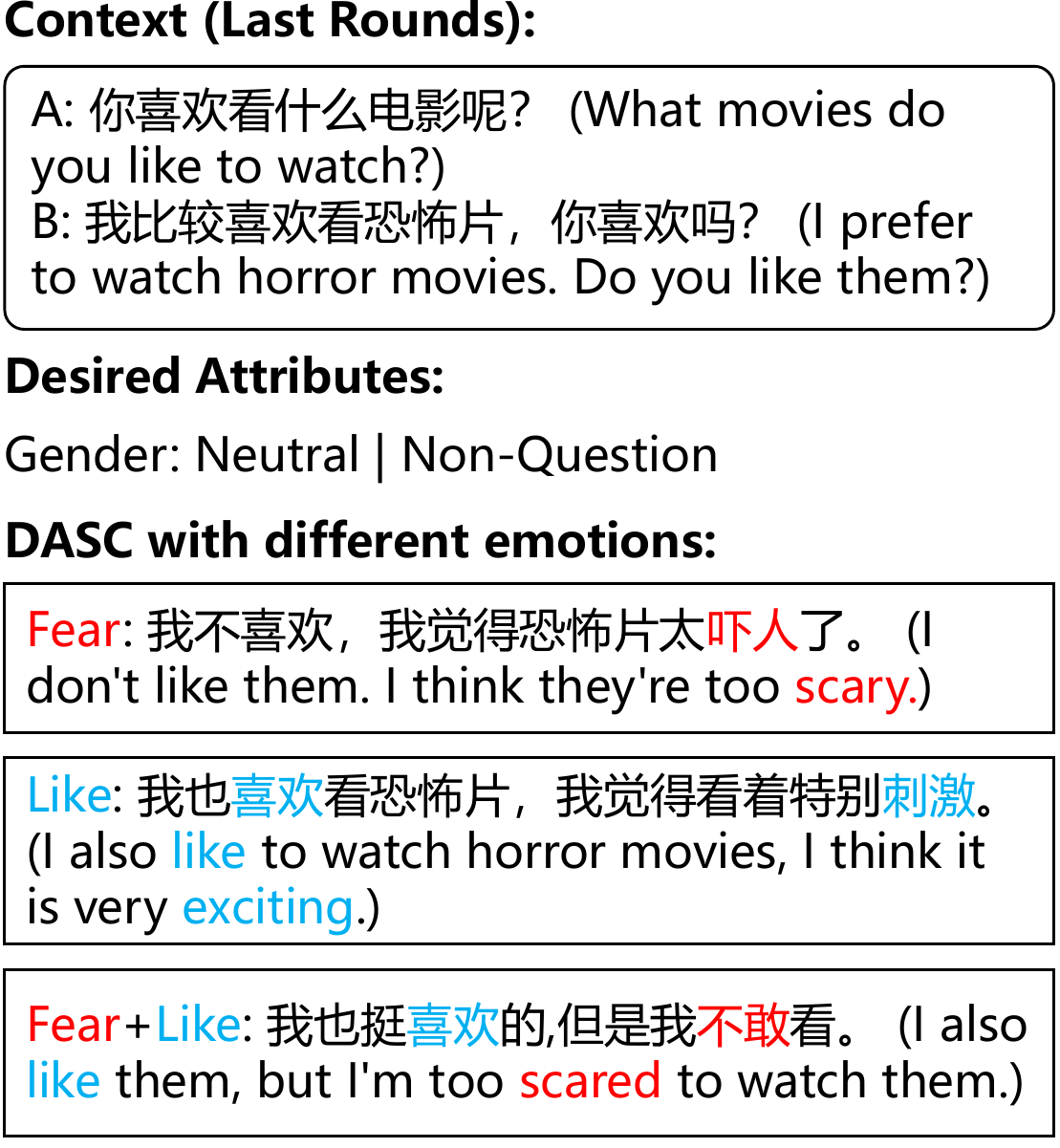}
    \caption{Example of DASC composing \textit{Like} and \textit{Fear} 
emotion in the generated response.}
    \label{fig:compose_example2}
\end{figure}

Figure \ref{fig:compose_example1_zh} shows the original Chinese text for the emotion composition example in Figure \ref{fig:compose_example1_en}, and we provide another example in Figure \ref{fig:compose_example2}, which shows that DASC can even compose a positive emotion \textit{Like} and a negative emotion \textit{Fear} in the same response to express complex meanings. 

\begin{figure}[ht]
    \centering
    \includegraphics[width=1.0\columnwidth]{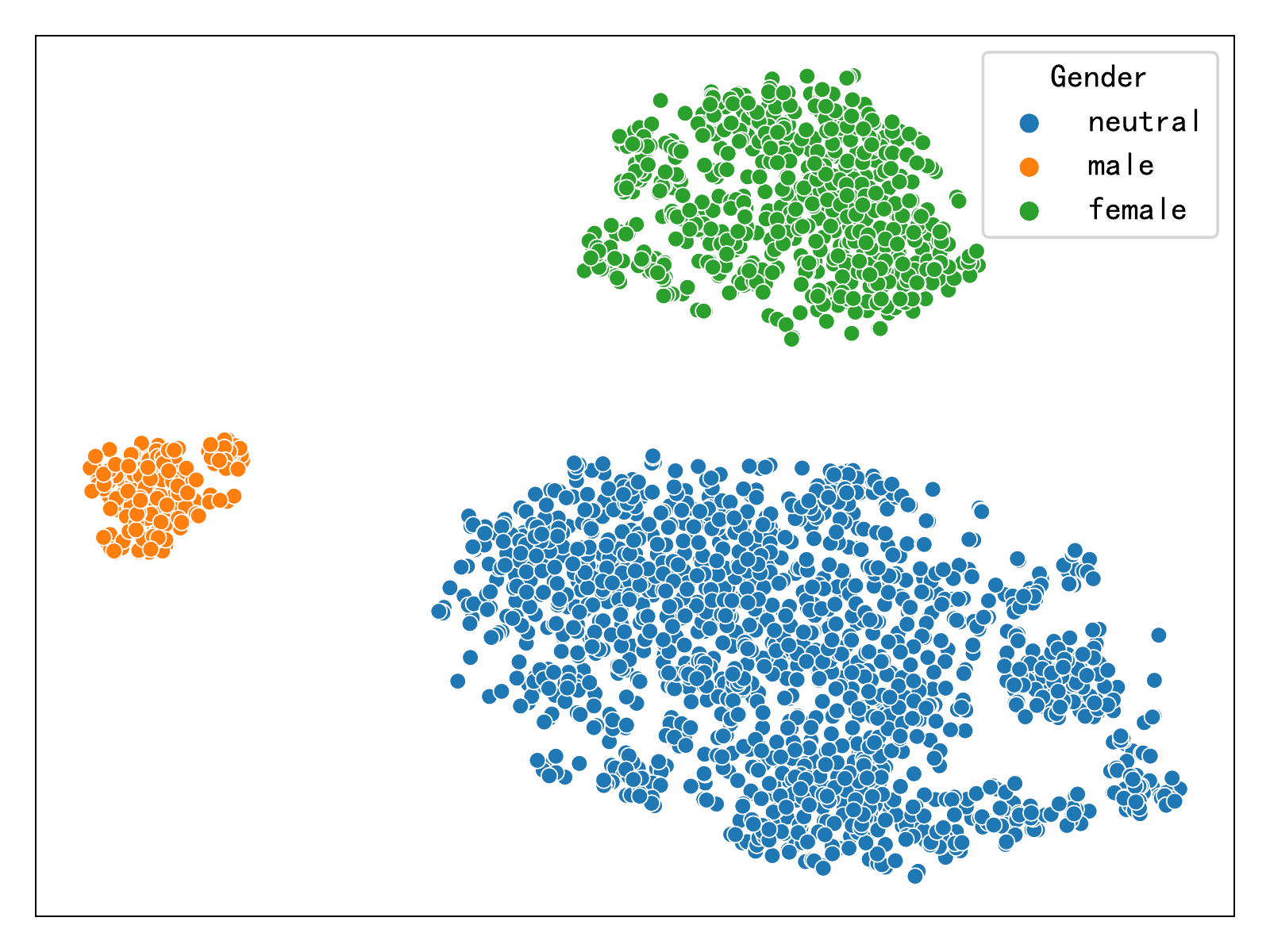}
    \caption{Visualization of attribute context embedding of responses with different gender styles.}
    \label{fig:gender_context_emb}
\end{figure}

\begin{figure}[ht]
    \centering
    \includegraphics[width=1.0\columnwidth]{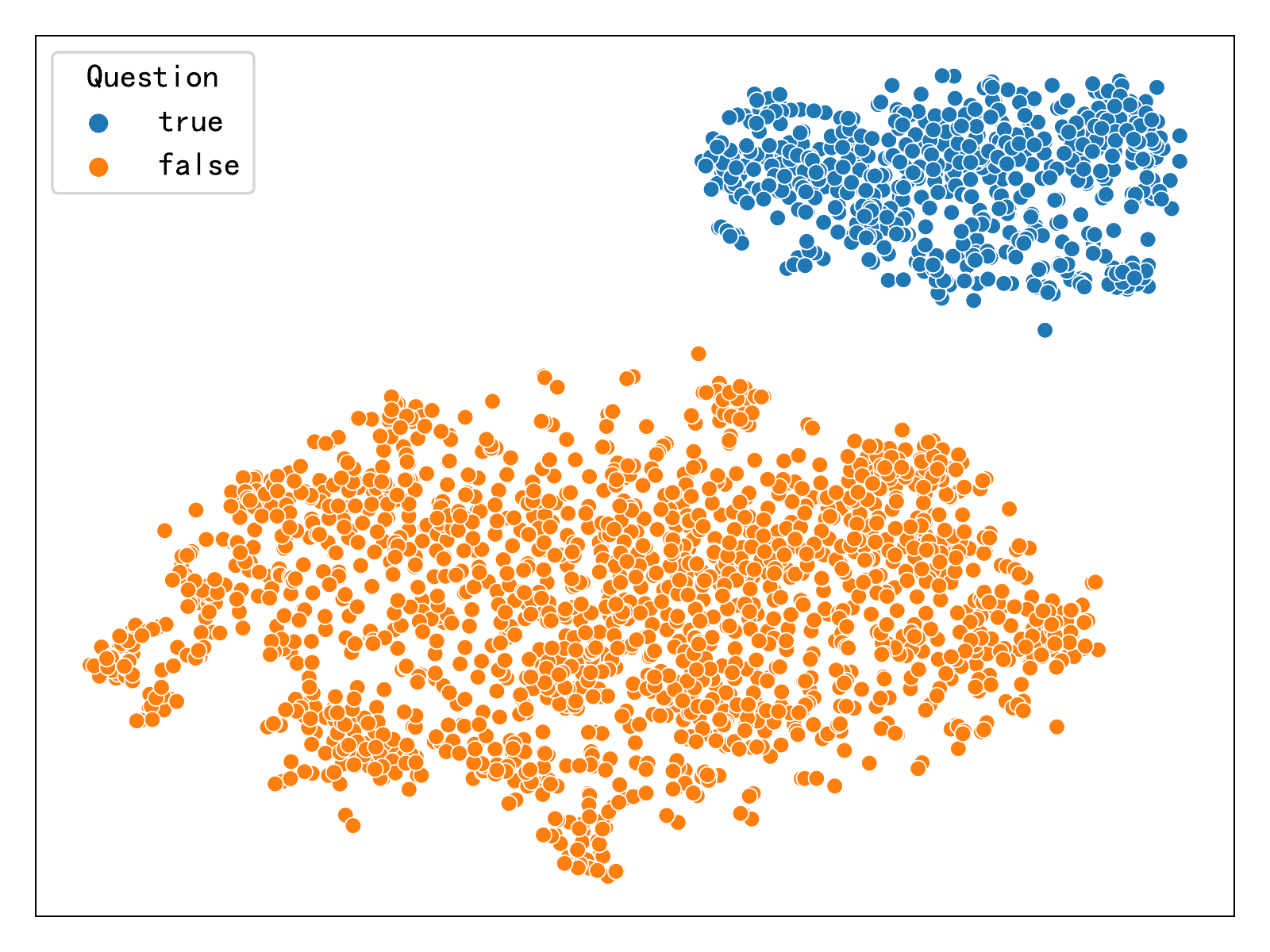}
    \caption{Visualization of attribute context embedding of responses with different question acts.}
    \label{fig:question_context_emb}
\end{figure}

We provide the t-SNE visualizations of attribute context embeddings of sentences with different \textit{Gender Style} and \textit{Question Act} in Figure \ref{fig:gender_context_emb} and Figure \ref{fig:question_context_emb}. We find similar results as we've seen for emotion, that the embeddings from different attributes are clearly separated.

\begin{figure}[htbp]
    \centering
    \includegraphics[width=1.0\columnwidth]{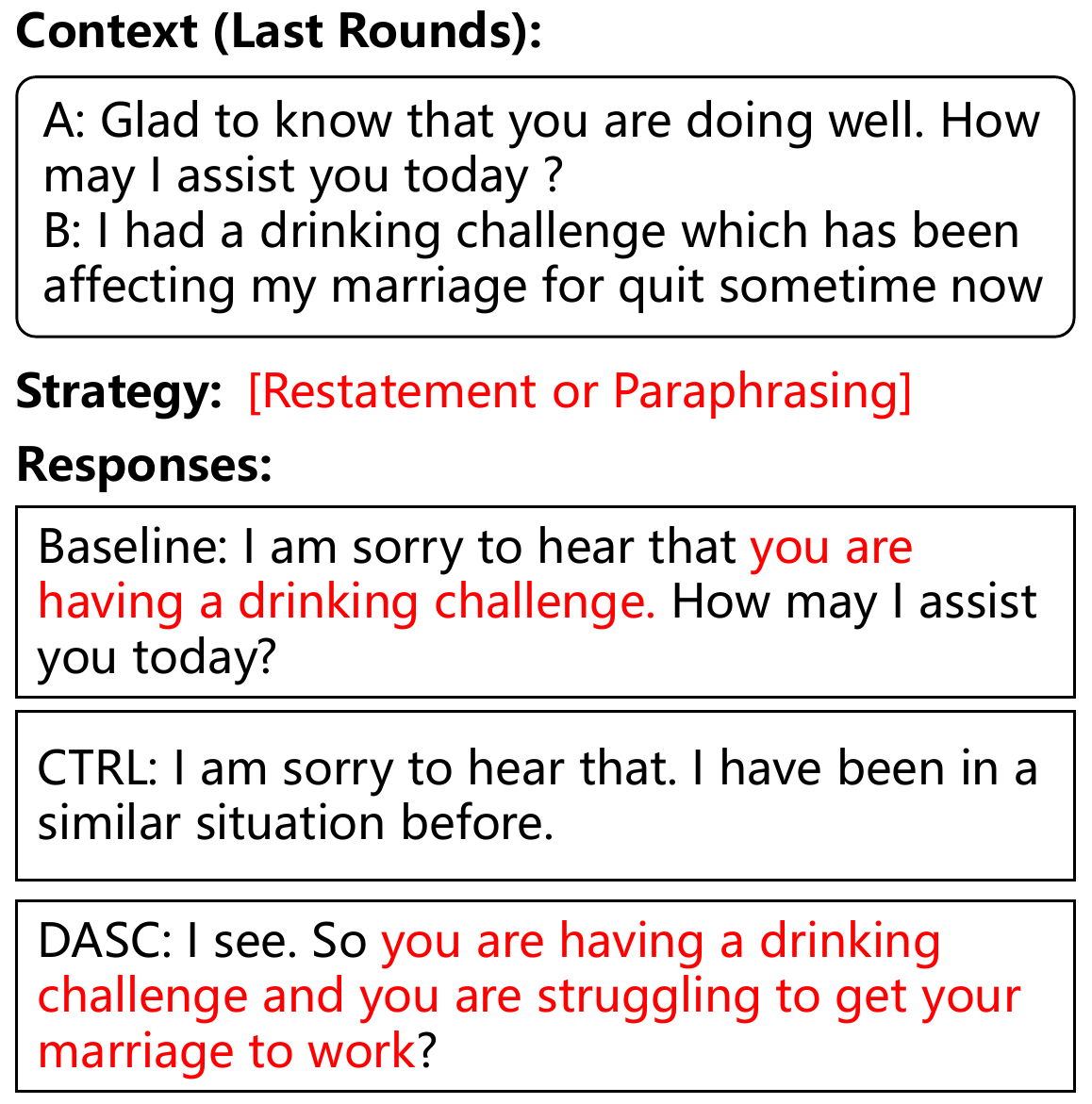}
    \caption{System generations in one example of ESConv, with the ``Restatement or Paraphrasing'' strategy.}
    \label{fig:esconv_example1}
\end{figure}

\begin{figure}[htbp]
    \centering
    \includegraphics[width=1.0\columnwidth]{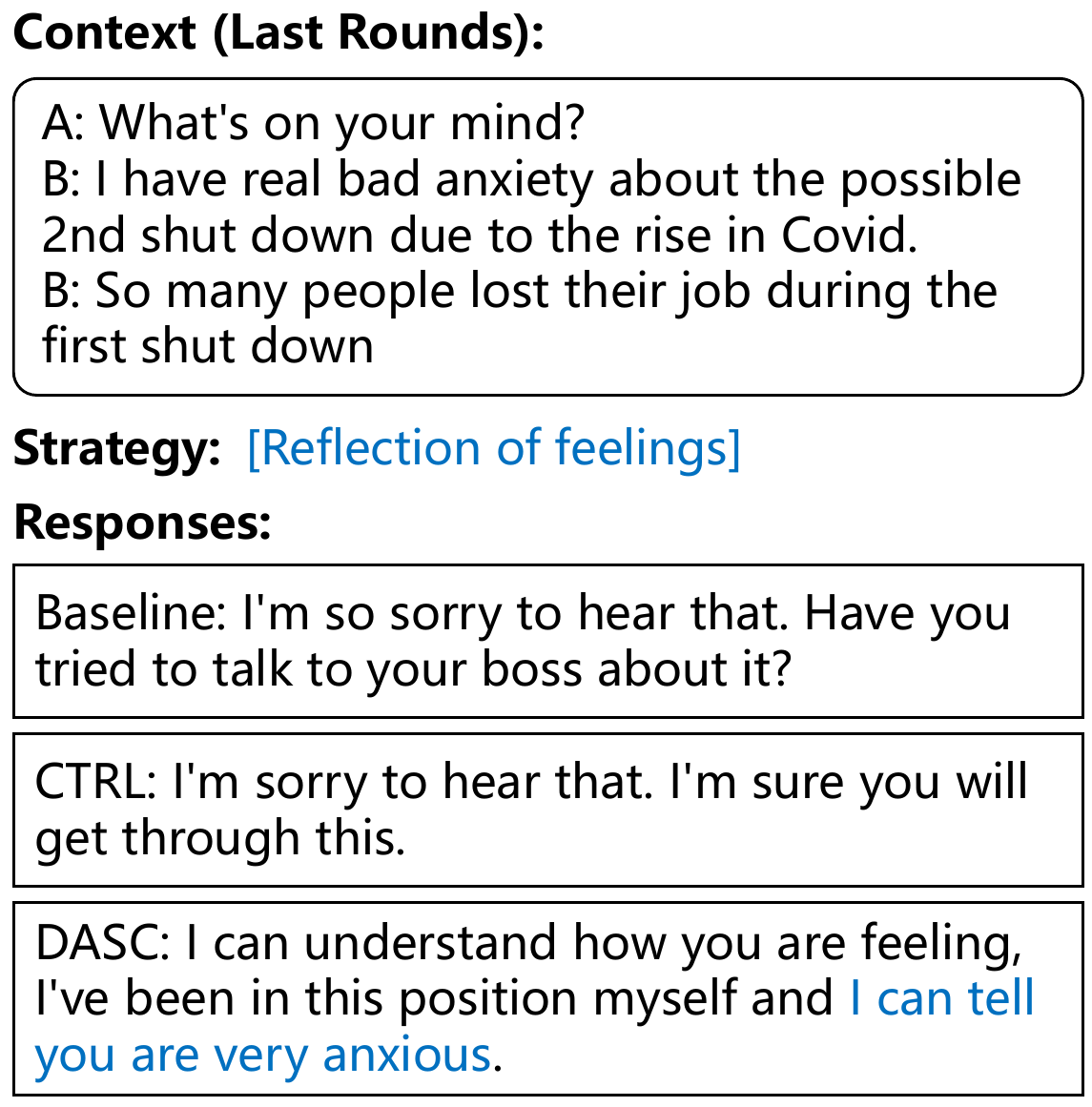}
    \caption{System generations in one example of ESConv, with the ``Reflection of feelings'' strategy.}
    \label{fig:esconv_example2}
\end{figure}

We also show 2 examples of the generated results on the ESConv. In Figure \ref{fig:esconv_example1}, both baseline and DASC successfully applied the desired strategy, while CTRL failed to do so. However, baseline also included a repetitive question at the end, while DASC gives a more comprehensive restatement, which exhibits a deep understanding of the situation and will be regarded as more helpful for the help seeker. In Figure \ref{fig:esconv_example2}, only DASC used the correct strategy in the generation, and such precise reflection of the anxious mood makes the response more sympathetic.

\end{document}